\documentclass[letterpaper, 10 pt, conference]{ieeeconf}  

\IEEEoverridecommandlockouts                              

\usepackage{amsmath}
\usepackage{amssymb}
\usepackage{bm}
\usepackage{booktabs}
\usepackage{graphicx}
\usepackage{wrapfig}
\usepackage{siunitx}
\usepackage{float}
\usepackage{hyperref}
\usepackage{float}
\usepackage{capt-of}
\usepackage{caption}
\usepackage{cite}
\newcommand{\web}{https://jessicayin.github.io/tactile-skin-rl}
\newcommand{\secv}{\vspace{-0.5em}}

\title{\LARGE \bf
Learning In-Hand Translation Using Tactile Skin With Shear and Normal Force Sensing
}

\author{Jessica Yin$^{1,2}$, Haozhi Qi$^{1,3}$, Jitendra Malik$^{1,3}$, James Pikul$^{4}$, Mark Yim$^{2}$, and Tess Hellebrekers$^{1}$
\thanks{$^{1}$Meta AI Research in
        Redmond, Washington, USA }%
\thanks{$^{2}$University of Pennsylvania GRASP Lab in
        Philadelphia, Pennsylvania, USA}
\thanks{$^{3}$UC Berkeley in
        Berkeley, California, USA}
\thanks{$^{4}$University of Wisconsin-Madison in
        Madison, Wisconsin, USA}
\thanks{This work was conducted while Jessica Yin was an intern and Haozhi Qi was a research fellow with Meta FAIR. This work was also supported in part by ONR MURI N0001421-1-2801, NSF GRFP grant 202095381, and NSF grant 1935294.}
}

\begin{document}

\maketitle
\thispagestyle{empty}
\pagestyle{empty}

\begin{abstract}

Recent progress in reinforcement learning (RL) and tactile sensing has significantly advanced dexterous manipulation. However, these methods often utilize simplified tactile signals due to the gap between tactile simulation and the real world. We introduce a sensor model for tactile skin that enables \textit{zero-shot} sim-to-real transfer of ternary \textit{shear} and binary \textit{normal forces}. Using this model, we develop an RL policy that leverages sliding contact for dexterous in-hand translation. We conduct extensive real-world experiments to assess how tactile sensing facilitates policy adaptation to various unseen object properties and robot hand orientations. We demonstrate that our 3-axis tactile policies consistently outperform baselines that use only shear forces, only normal forces, or only proprioception. Videos and details available on the \href{\web}{project website}.

\end{abstract}


\section{Introduction}

Humans rely on their sense of touch to manipulate objects in their daily lives~\cite{johansson2009coding}. Inspired by this, a prominent area of manipulation research focuses on equipping robots with tactile sensors~\cite{lee2000tactile,howe1993tactile}. However, despite the development of capable tactile sensors~\cite{yuan2017gelsight,lambeta2020digit,bhirangi2021reskin}, there remains a significant gap between the breadth of information they can capture and what state-of-the-art robot controllers can utilize for feedback. Recently, reinforcement learning (RL) and sim-to-real techniques have enabled breakthroughs in integrating tactile feedback into low-level controllers for dexterous manipulation~\cite{qi2023general,yin2023rotating,khandate2023sampling}. However, these are constrained by the speed and accuracy of tactile simulation techniques and often rely on simplified tactile signals. Although tactile simulation has been actively pursued recently~\cite{xu2023efficient,wang2022tacto,si2024difftactile}, simulating rich, large-area, and high-fidelity tactile sensing with high throughput remains a challenge.

Tactile sensors approximate contact through soft body deformation, which is traditionally modeled using slow and high-fidelity techniques like FEM~\cite{ma2019dense,sferrazza2019ground,du2024tacipc}. At the other extreme, most simulators fast enough to train RL controllers use overly simplified contact models for numerical stability, which struggle to bridge the sim-to-real gap. Robotic dexterity requires a balance between simulation speed and fidelity to fully leverage the progress in both tactile sensing and RL techniques. To that end, we introduce a tractable tactile skin model that outputs normal and shear forces to enable policy training at 5000 FPS on two GPUs, achieving 70\% of training speed without touch. We demonstrate the fidelity of the proposed tactile sensor model through zero-shot sim-to-real transfer of policies trained entirely in simulation.

To investigate whether our ternary shear (\{-1, 0, 1\}) and binary normal (\{0, 1\}) forces improve policy performance, we specifically study in-hand translation of object across the palm. With current limitations in simulating contact dynamics and tactile skin, we have yet to see examples of this dexterous skill with tactile feedback including shear forces. In-hand translation is especially interesting for shear feedback because it requires controlled sliding of the objects~\cite{cole1992dynamic,yang2024dynamic,teeple2022controlling}, and this skill is useful for practical applications such as repositioning tools to a functional grasp.


\begin{figure}[!t]
\centering
\includegraphics[width=\columnwidth]{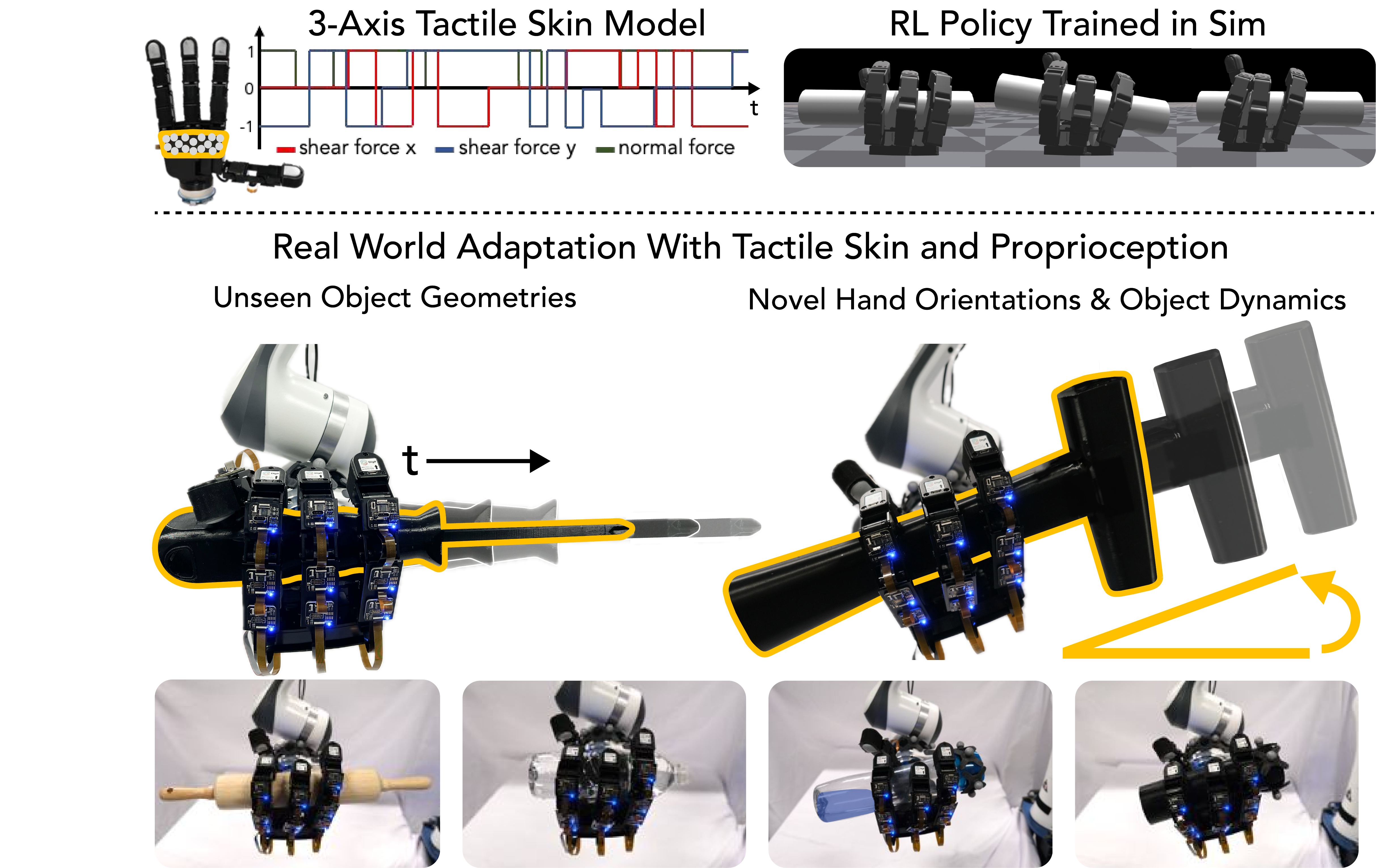}
\vspace{-1em}
\caption{\small We present a tactile skin model that enables \textit{zero-shot} sim-to-real transfer of ternary shear and binary normal forces. We use it to learn an RL policy for dexterous in-hand translation that uses tactile and proprioceptive feedback to adapt to unseen object geometries, novel hand orientations, and new object dynamics. We evaluate our policies with over 190 real-world rollouts, available \href{\web}{online}.}
\vspace{-2em}
\label{fig:overview}
\end{figure}
In summary, we introduce our sim-to-real approach for in-hand object translation using tactile sensing. We make three key contributions: 
\begin{itemize}
    \item To the best of our knowledge, we contribute the first RL-tractable sensor model for compliant tactile skin that enables \textit{zero-shot} sim-to-real transfer of \textbf{binary normal \textit{and ternary shear}} signals.
    \item We use our three-axis sensor model to learn RL control policies for in-hand translation, a dexterous, contact-rich task that requires controlled sliding contact.
    \item We show that our control policies with three-axis tactile sensing achieve superior in-domain task performance and adaptation to unseen objects and hand orientation, with 190 \textit{real-world} rollouts.
\end{itemize}

\section{Related Work}

\subsection{Tactile Sensor Simulation} 

The crux of tactile sensor simulation is efficiently and sufficiently modeling the deformations and forces at the soft contact interface. Most works focus on simulating fingertip visuo-tactile sensors, which use cameras to observe soft material deformations upon contact \cite{xu2023efficient,lin2022tactile,chen2023tacchi}. Finite Element Method (FEM)-based approaches are high fidelity \cite{si2022taxim,narang2021sim,luu2023simulation}, but their computational cost prevents their use in RL policy learning. \cite{si2024difftactile} uses FEM to demonstrate a sim-to-real RL policy for two-fingered grasping, but the tractability may not extend beyond this small action space. Tacto \cite{wang2022tacto}, Mujoco \cite{todorov2012mujoco}, and IsaacGym \cite{makoviychuk2021isaac} use collision geometry penetration to efficiently estimate forces. While this can be sufficiently accurate for normal forces, it produces sparse signals for shear. \cite{xu2023efficient} leverages rigid-body assumptions in tactile modeling for sim-to-real, but the only task transferred to the real world assumes constant and sticking contact, and other tasks with diverse contact modes are not demonstrated.

Far fewer recent works explore non-camera based tactile skin simulation \cite{cremer2020skinsim, kappassov2020simulation}. These approaches decouple deformation modeling and sensor response to account for cross-talk and noise, producing realistic data but suffering from intractability for RL policy training. \cite{yin2023rotating, yuan2023robot} simulate force-sensitive resistors as binary tactile sensors using the default IsaacGym tactile sensor model. In contrast, our work introduces a simulation model for both normal and shear forces for a magnetic tactile skin~\cite{bhirangi2021reskin, hellebrekers2019soft}.

\subsection{In-Hand Manipulation} 

In-hand manipulation has been studied for decades using either classical control~\cite{teeple2022controlling,morgan2022complex,han1998dextrous,saut2007dexterous,rus1999hand,bai2014dexterous,mordatch2012contact,fearing1986implementing} approaches or learning-based methods. 
Learning-based methods can be categorized to real-world imitation learning~\cite{arunachalam2023holo,haldar2023teach,arunachalam2023dexterous,qin2022one,wang2024dexcap} and sim-to-real with reinforcement learning~\cite{openai2018learning,openai2019solving,nvidia2022dextreme,chen2023visual,lin2024twisting}. Our method falls into the latter category and differs from existing work from two perspectives. First, the majority of existing methods focus on in-hand object reorientation~\cite{qi2023general,yin2023rotating,qi2022hand,yang2024anyrotate} while our work focuses on in-hand object translation. This task is a complementary skill with in-hand rotation, and a critical step towards general in-hand manipulation. Second, most existing learning-based in-hand manipulation systems use vision~\cite{chen2023visual} and proprioception~\cite{qi2022hand}, or simplified touch sensing limited to normal forces~\cite{qi2023general,yin2023rotating}. In contrast, our touch sensing pipeline provides both shear and normal forces for controller feedback.
\vspace{-1em}
\section{Tactile Skin Model}

\begin{figure}[!t]
    \centering
    \includegraphics[width=\columnwidth]{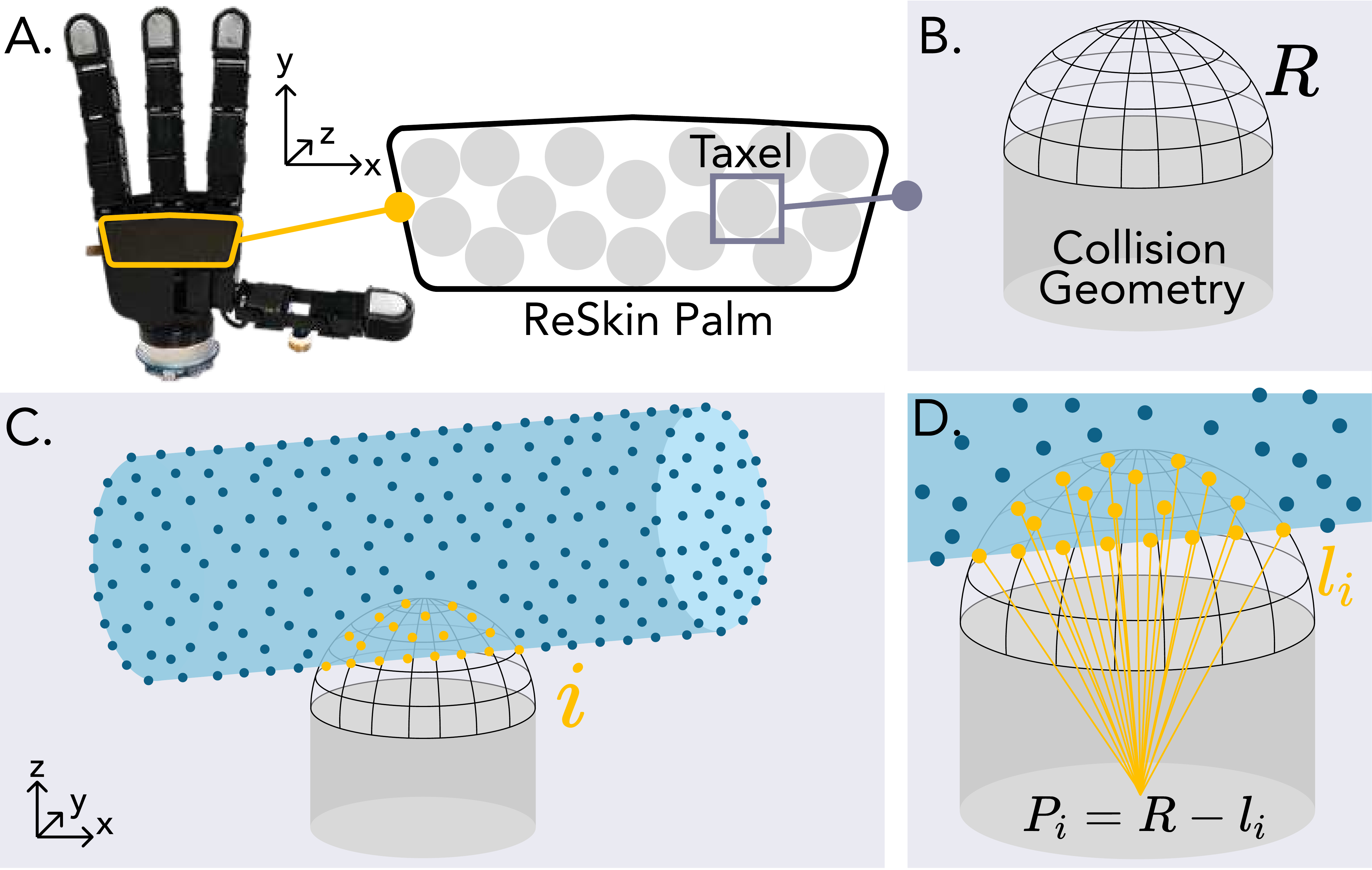}
    \vspace{-1.3em}
    \caption{\small \textbf{Our modeling approach for sim-to-real transfer of tactile skin.} A) We model the palm as a continuous surface with 16 discrete taxels, each corresponding to an underlying magnetometer. B) We use a cylinder for each taxel and extend the sensing range ($R$) beyond its collision geometry. C) We sample points on the object and represent the collision surface as a point cloud. Points within sensing range are denoted as $i$. D) We sum the penetration distances $P_i = R - l_i$, where $l_i$ is the distance from point $i$ to the sensor's origin. We calculate the sensor signals for shear and normal force using $\sum_{i=1}^{n} P_{i}$, object velocity, and object point density.}
    \label{fig:sensor_model}
    \vspace{-1.6em}
\end{figure}
Our sensor model focuses on ReSkin \cite{bhirangi2021reskin}, a magnetic elastomer tactile skin, which we adhere to the palm of the robot hand (Figure \ref{fig:sensor_model}A). As the skin deforms and the positions of the embedded magnetic particles change, underlying magnetometers measure the magnetic flux as a proxy for 3-axis force. To minimize the effects of hysteresis and cross-talk, we binarize the sensor outputs in both simulation and reality. We implement our tactile sensor model and train our RL policy in IsaacGym~\cite{makoviychuk2021isaac}, and it achieves around 70\% of training speed compared to training without any tactile sensors. For context, with 2 NVIDIA RTX-4090 GPUs, we get 5000 FPS training with tactile sensors and approximately 7000 FPS without tactile sensors.

\begin{figure}[!t]
\centering
\includegraphics[width=\columnwidth]{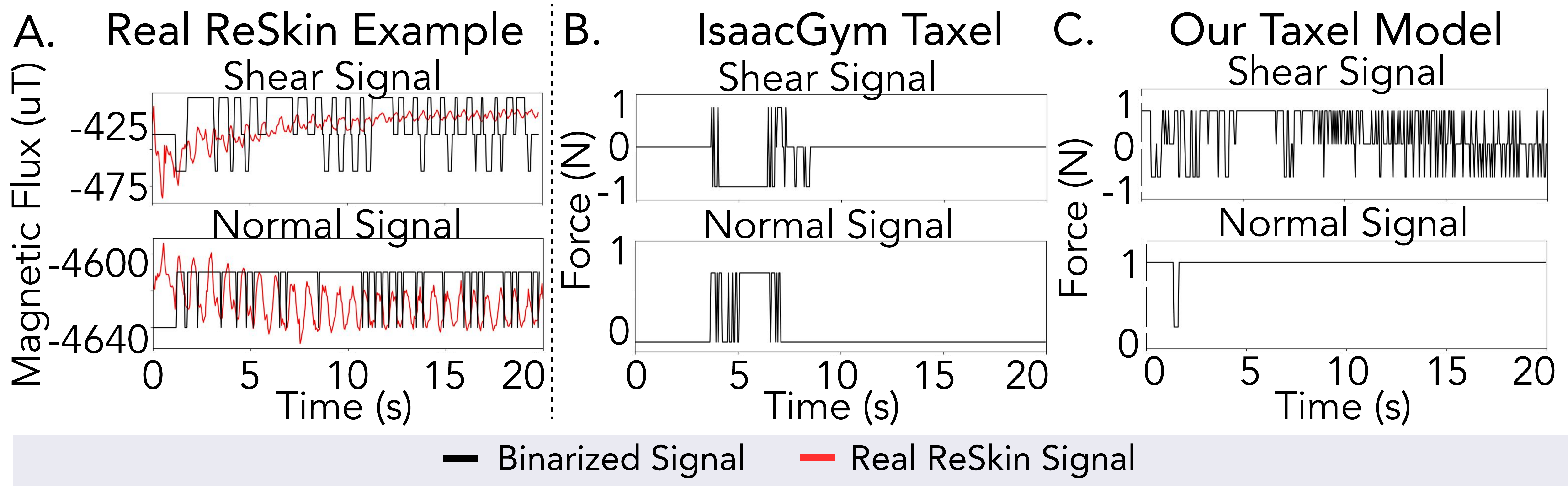}
\vspace{-1.5em}
\caption{\small \textbf{Tactile signal examples during object translation. This is a \textit{representative example, not a direct comparison, of simulation and reality} due to the differences in object trajectories during the task.} A) An example of \textbf{real-world} ReSkin palm taxel output. The signals are periodic because the finger gait is periodic. B) The output of a \textbf{simulated} palm taxel, using the default IsaacGym force sensor similar to \cite{yin2023rotating, yang2024anyrotate}. The signals are sparse because the model relies on collision geometry penetration. C) The taxel outputs from our \textbf{simulated} S3-Axis tactile skin model, for the \textit{same taxel} and the \textit{same rollout} as B.}
\label{fig:reskin_comparison}
\vspace{-2em}
\end{figure}

Our approach contrasts with the default tactile sensor models offered in simulators such as Mujoco and IsaacGym, which report normal and shear signals calculated primarily from collision geometry penetration (Figure \ref{fig:reskin_comparison}). Although this approach offers numerical stability in simulation by neglecting rapid changes in frictional forces, it produces sparse and unrealistic signals for shear forces in tactile sensor models. Our sensor model outputs sufficiently accurate shear and normal forces for sim-to-real transfer, while still leveraging the speed and stability of the simulator by using collision geometry penetration to calculate dynamics and contact interactions.

\textbf{Discretization.}
The ReSkin palm is modeled in two parts. First, we model the ReSkin palm pad as a rigid volume with the same physical dimensions as the real sensing skin. The simulated ReSkin palm pad acts as a continuous collision surface for more accurate dynamics during the task. Second, although ReSkin is continuous, we discretize the sensing skin to 16 discrete taxels, each corresponding to the location of an underlying magnetometer (Figure \ref{fig:sensor_model}A). We model each ReSkin taxel as a cylinder collision geometry ($r=1.5$ cm), placed coincidentally on the surface of the ReSkin palm.

\textbf{Shear and Normal Sensor Signals.}
To calculate shear signals, we draw inspiration from compliant contact models \cite{lynch2017modern, le2023differentiable, drake}. We use: $S_{x,t} = \frac{\sum_{i=1}^{n} P_{i}}{D}(v_{x,t} + \omega_{x,t})$ and $S_{y,t} = \frac{\sum_{i=1}^{n} P_{i}}{D}(v_{y,t} + \omega_{y,t})$ where $x$ and $y$ correspond to global coordinate axes and are tangential to the sensor surface, $S_{x,t}$ and $S_{y,t}$ are the shear signals produced by each taxel, $v_{x,t}$ and $v_{y,t}$ are the linear velocities of the object, $\omega_{x,t}$ and $\omega_{y,t}$ are the angular velocities of the object, $\sum_{i=1}^{n} P_{i}$ is the summation of object point penetration distances within the sensing range, and $D$ is the object point density (total number of object points divided by object volume). The point penetration distance is the $L_2$ norm of sensor range and the 3D object point coordinates. For the normal force signal, we use $S_{z,t} = \frac{\sum_{i=1}^{n} P_{i}}{D}$. We calculate the normal force with the summation of object point penetration distances, divided by object point density. The object point penetration distances are a proxy for normal force. Dividing by object point density accounts for different scales of the object.

To bridge the reality gap, we train the control policy with thresholded sensor signals from our tactile skin model: $S_{x,t} \in \{-1, 0, 1\}, S_{y,t} \in \{-1, 0, 1\}, \text{ and } S_{z,t} \in \{0, 1\}$. For the signed 3-axis policy variant (\textit{S3-Axis}), we retain positive and negative signs of shear signals, which give direction along each axis. For the unsigned 3-axis policy variant (\textit{U3-Axis}), we take the absolute value of the shear signal. 


\textbf{Real-World Binary Tactile Processing.} 
Raw signals from ReSkin have significant hysteresis, such that simple thresholding is not sufficient for binarization. Instead, we use a threshold on the \textit{time derivative} of the signal. We use two buffers: one for signal history and one for the current signal. If the difference between the signal history and current signal buffers exceeds the threshold, then a $1$ or $-1$ is returned (else $0$). We tune buffer sizes and thresholds per $x$, $y$, $z$ axis. There is a slight delay in binary sensor outputs due to the dependence on signal history, but because ReSkin outputs data four times faster (\SI{78}{\hertz}) than the policy sampling rate (\SI{20}{\hertz}), we find this delay to be negligible. 


\section{In-Hand Translation with Tactile Skin}

\begin{figure}[!t]
\centering
\includegraphics[width=\linewidth]{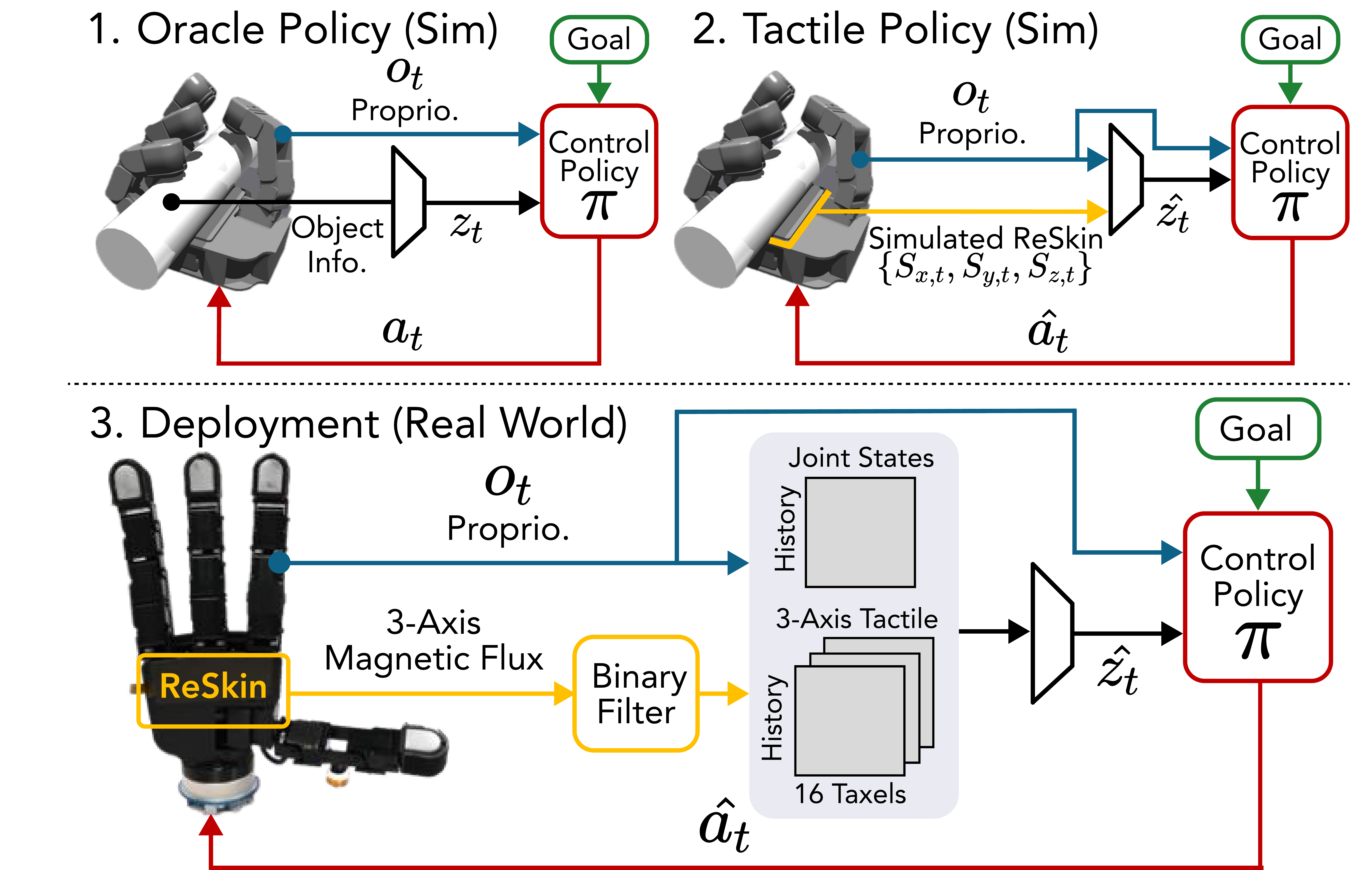}
\vspace{-1em}
\caption{\small \textbf{Overview of our training and deployment pipeline.} We train the policy in two stages, entirely in simulation. We use our tactile skin model in the second stage. The policy is directly deployed in the real world.}
\vspace{-2em}
\label{fig:system}
\end{figure}

To demonstrate the effectiveness of our simulated tactile skin model, we perform sim-to-real experiments on the in-hand translation task. This task requires rich, controlled sliding contact on the palm, so we hypothesize that palm tactile feedback can be particularly useful.

\textbf{Policy Pipeline.} We adopt a two-stage method for policy learning from~\cite{qi2022hand,qi2023general}. First, we train an oracle policy $\pi$ with privileged information openly available from the simulator. Second, we train the tactile policy with an observation encoder using our simulated sensor model and freeze the control policy $\boldsymbol{\pi}$. During deployment, we directly deploy the control policy and observation encoder in the real world. An overview is shown in Figure~\ref{fig:system}.

\textbf{Inputs and Outputs.} The privileged information includes object state and physical properties sampled from the simulator. This information includes object position, velocity, size, mass, center of mass, and coefficient of friction. It is then encoded in an 8-dimensional vector, \boldsymbol{$z_t$}, representing information that the policy finds useful and relevant for the task. The inputs to the oracle policy, \boldsymbol{$\pi$}, are finger joint positions from proprioception and the encoded privileged information, \boldsymbol{$z_t$}. The policy outputs the 16 joint position targets of the PD controller, \boldsymbol{$a_t$} $\in \mathbb{R}^{16}$. The observation \boldsymbol{$o_t$} contains a temporal window of joint positions and actions, \boldsymbol{$o_t$} $=$ $[$\boldsymbol{$q_t$}, \boldsymbol{$a_t$}$]$ $\in \mathbb{R}^{96}$. This gives us \boldsymbol{$a_t$} $=$ \boldsymbol{$\pi$}$($\boldsymbol{$o_t$}$,$ \boldsymbol{$z_t$}$)$.

\textbf{Task Training. }The task requires the control policy to translate objects in one direction across the palm. Our reward function uses penalties on object state to specify the in-hand translation task and penalties on robot behavior to produce finger gaiting. The reward contains three main terms: task rewards ($r_{\text{iht}}, r_{\text{goal}}$) to define the task, motion penalties ($r_{\text{rotp}}, r_{\text{pose}}$), and energy penalties ($r_{\text{work}}, r_{\text{torque}}, r_{\text{force}}$) (Table \ref{table:reward}). We use PPO \cite{schulman2017proximal} to optimize the oracle policy and share weights between the policy and the critic network. An extra linear projection layer estimates the value function. During training, a cylinder with randomized physical properties is initialized in a stable grasp for each environment and we assign a random goal position to the environment.

The key difference between the oracle policy and the tactile policy is the observation encoder. We concatenate simulated sensor data from proprioception (\boldsymbol{$q_t$}) and our tactile skin model ([\boldsymbol{$S_{x,t}, S_{y,t}, S_{z,t}$}]) as inputs to the observation encoder. The observation encoder is a transformer trained to minimize the $L_2$ norm between: 1) \boldsymbol{$z_t$} and \boldsymbol{$\hat{z}_t$}, aiming to replicate the privileged representation from simulated sensor data, and 2) \boldsymbol{$a_t$} and \boldsymbol{$\hat{a}_t$}, aiming to replicate the same actions as the oracle control policy.
\begin{table}[H]
\centering
\setlength{\tabcolsep}{4pt}
\renewcommand{\arraystretch}{1.1}
\resizebox{\linewidth}{!}{%
\begin{tabular}{cc}
\toprule  
\textbf{Reward} & \textbf{Scale}\\
\midrule
$r_{\text{iht}} \doteq -(x_{\text{obj\_pos}} - x_{\text{obj\_goal\_pos}})^2$ &$700$ \\
$r_\text{rotp} \doteq -(x_\text{obj\_left\_end} - x_\text{obj\_right\_end})^2$ &   $500$    \\
$r_\text{goal} \doteq (1  \, \text{if} \, \sqrt{(x_{\text{obj\_pos}} - x_{\text{obj\_goal\_pos}})^2} < \epsilon \, \text{else} \, 0)$  &  $10$     \\
$r_\text{drop} \doteq \min(\max((x_\text{obj\_pos} - x_\text{threshold}), -1), 0)$ &  $1000$      \\
$r_{\text{pose}} \doteq - ||$\boldsymbol{$q$} $-$ \boldsymbol{$q_{\text{init}}$}$||^2_2$  &  $-0.3$        \\
$r_{\text{work}} \doteq -\boldsymbol{\tau}^T \boldsymbol{\dot{q}}$  &  $-2.0$ \\
$r_{\text{torque}} \doteq -||$\boldsymbol{$\tau$}$||^2_2$ &  $-0.1$ \\
$r_\text{force} \doteq -\frac{1}{4} \sum_{i=1}^{4} (F_i - \mu)^2$ &  $500$\\
\bottomrule
\end{tabular}
}
\vspace{-0.3em}
\captionof{table}{\normalfont Reward function for the oracle policy.}
\label{table:reward}
\vspace{-1em}
\end{table}
\section{Experiment Setup}

\textbf{Hardware.} We use an Allegro Hand~\cite{allegro} for our experiments. It has four fingers, each with four degrees of freedom (DoF). The  16 joints receives target joint position from neural network controller at \SI{20}{\hertz}, and the commands will be converted to torque using a PD controller at \SI{300}{\hertz}. ReSkin, measuring approximately 37 mm x 96 mm, is adhered to the palm of the Allegro Hand and outputs tactile data in $\mathbb{R}^{16 \times 3}$ at \SI{78}{\hertz}. Since there is high friction between the objects and ReSkin, the ReSkin palm is covered with a 0.25 mm sheet of PET-like plastic to reduce the required torque from the robot fingers for the task, thus avoiding overheating.

\begin{figure}[!t]
\centering
\includegraphics[width=\linewidth]{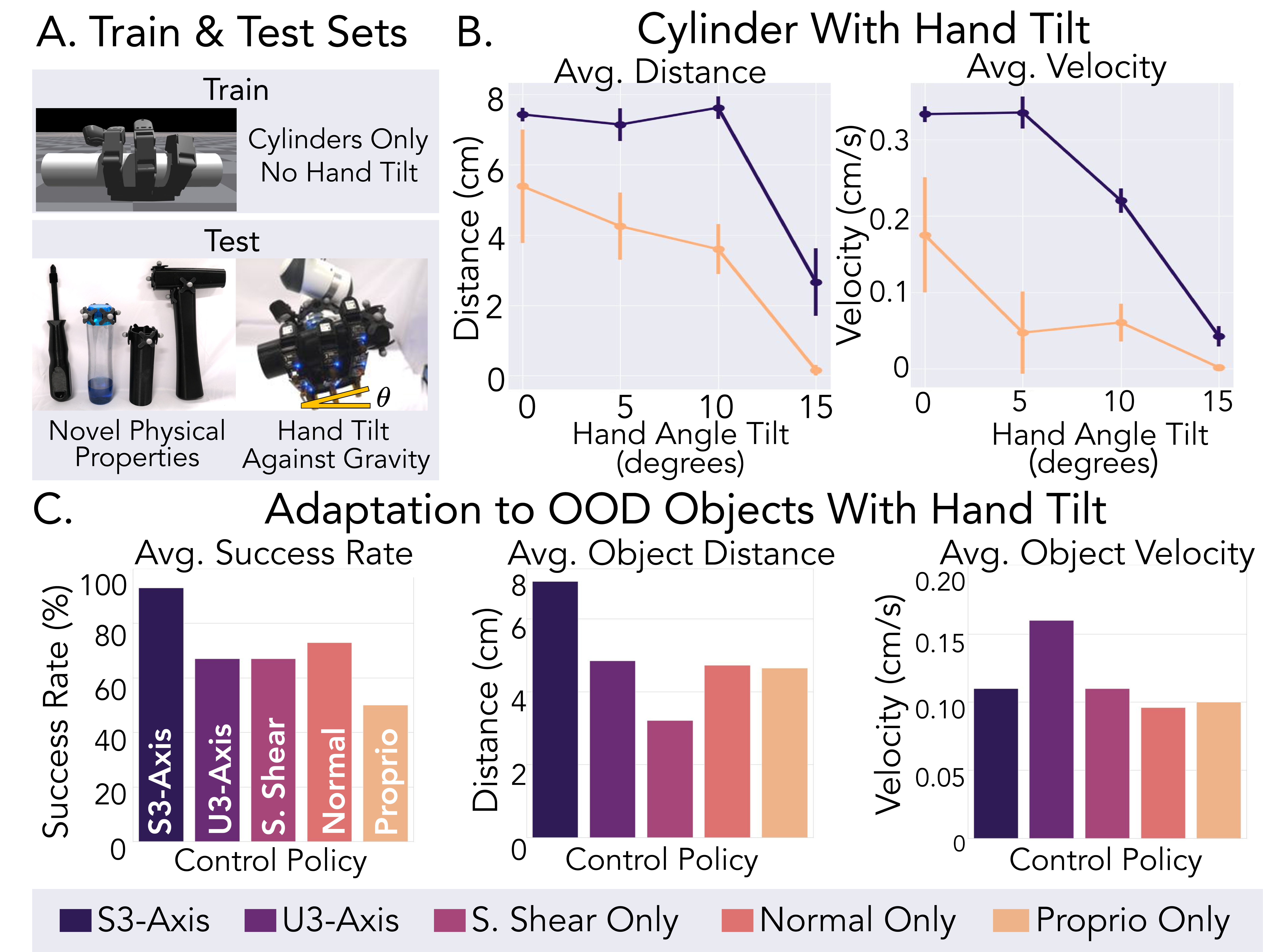}
\vspace{-1em}
\caption{\small \textbf{A. Train and test sets.} We train with cylinders and no hand tilt in simulation. We test on real objects with varying COM, geometries, and hand angles. Motion capture markers on the objects are only for measuring task metrics. \textbf{B. Real-world cylinder rollouts with S3-Axis and Proprio-Only}. This shows superior S3-Axis policy performance compared to Proprio-Only for both ID and OOD conditions. Error bars indicate standard deviation. \textbf{C. Three-axis tactile sensing policies demonstrate the best adaptation to OOD objects and unseen hand orientations.} S3-axis enables \textbf{93\% average success rate} and \textbf{+51\% increase} in distance over Proprio-Only. U3-axis enables \textbf{+60\% increase} in velocity over Proprio-Only. These metrics are averaged over all real-world OOD experiments (30 rollouts/policy).}
\label{fig:avg_adapt}
\vspace{-2em}
\end{figure}

\textbf{Simulation.} We use the IsaacGym~\cite{makoviychuk2021isaac} simulator to train our policy. Each environment contains a robot hand and randomly scaled cylinder (Figure~\ref{fig:avg_adapt}A). We use our model from Section to simulate tactile data. The simulation frequency is \SI{200}{\hertz}, and the control frequency is \SI{10}{\hertz}. Each episode lasts 400 time steps (\SI{20}{\second}) and resets when the object falls.



\textbf{Ablations.} We ablate tactile modalities to compare our methods, \textit{Signed 3-Axis (S3-Axis)} and \textit{Unsigned 3-Axis (U3-Axis)}. All policies are trained with the same oracle policy, and all tactile policies use proprioception and 16 palm taxels.
    
\textit{Signed Shear Only.} The tactile sensors only output ternary values for shear forces: $S_{x,t}, S_{y,t} \in \{-1, 0, 1\}$ and $S_{z,t} \in {0}.$ 
    
\textit{Normal Only.} The tactile sensors only output binary values for normal forces: $S_{x,t}, S_{y,t} \in \{0\}$ and $S_{z,t} \in \{0, 1\}.$ This baseline is similar to \cite{yin2023rotating, yuan2023robot}.
    
\textit{Proprio Only.} The policy is not trained with tactile sensors.

\textbf{Metrics.} We use the following metrics to measure policy performance. We use Optitrack motion capture to track object pose at \SI{240}{\hertz} to calculate these metrics. For all metrics, higher is better. 
    
\textit{Success Rate.} Success is defined as the policy achieving an object translation distance greater than 0 mm within 120 s. We set the threshold to be 0 mm because some policies completely fail to move the object or drop it, especially when the hand is tilted.
    
\textit{Average Object Distance.} The average maximum distance (cm) an object moves along the desired translation axis, calculated from successful rollouts.
    
\textit{Average Object Velocity.} The average object velocity (cm/s) along the desired translation axis, calculated from successful rollouts.

\textbf{Object Dataset.} We evaluate policies with the following in-domain (ID) and out-of-domain (OOD) objects with differing centers of mass (COM) (Figure \ref{fig:avg_adapt}A). Our test set consists of real objects only: cylinder (ID, $6.5$ cm diameter x $22.2$ cm length, $108$ g), hammer (OOD, skewed COM, $37$ cm x $6.4$ cm x $20$ cm, $284$ g), screwdriver (OOD, challenging geometry, $3.8$ cm x $5.4$ cm x $39.4$ cm, $180$ g), and water bottle (OOD, variable center of mass, between $6$-$7$ cm, total weight $252$ g with $191$ g water). The hammer \cite{hammer} and screwdriver \cite{screwdriver} are scaled by 200\% to match the scale of an Allegro Hand compared to a human hand.
\vspace{-0.3em}
\section{Results and Analysis}

\begin{table}
\footnotesize
\centering
\begin{tabular*}{\columnwidth}{@{\extracolsep{\fill}}llccc}
\toprule
\multicolumn{5}{c}{\textbf{Screwdriver - Challenging Geometry}}  \\   
Hand Tilt & Policy     & \multicolumn{1}{l}{Success $\uparrow$} & \multicolumn{1}{l}{Dist. (cm) $\uparrow$} & \multicolumn{1}{l}{Vel. (cm/s) $\uparrow$} \\
\toprule
0 deg.          & S3-Axis & $100$    & $3.91\scriptstyle{\pm0.08}$      & $0.23\scriptstyle{\pm0.08}$\\
              & U3-Axis    & $100$     & $6.19\scriptstyle{\pm1.09}$       & $0.33\scriptstyle{\pm0.07}$\\
              & S. Shear Only          & $100$      & $3.22\scriptstyle{\pm2.5}$       & $0.14\scriptstyle{\pm0.29}$\\
              & Normal Only             & $100$      & $3.49\scriptstyle{\pm1.86}$      & $0.19\scriptstyle{\pm0.09}$\\
              & Proprio. Only & $100$        & $5.82\scriptstyle{\pm1.14}$       & $0.25\scriptstyle{\pm0.13}$\\
              \toprule
\multicolumn{5}{c}{\textbf{Hammer - Skewed COM}}  \\   
\toprule
15 deg.          & S3-Axis & $100$    & $12.53\scriptstyle{\pm1.08}$      & $0.23\scriptstyle{\pm0.21}$\\
              & U3-Axis     & $100$     & $5.30\scriptstyle{\pm4.40}$       & $0.05\scriptstyle{\pm0.11}$\\
              & S. Shear Only          & $100$      & $6.55\scriptstyle{\pm1.35}$       & $0.25\scriptstyle{\pm0.05}$\\
              & Normal Only             & $100$      & $7.36\scriptstyle{\pm5.93}$      & $0.14\scriptstyle{\pm0.16}$\\
              & Proprio. Only & $100$        & $11.33\scriptstyle{\pm1.02}$       & $0.23\scriptstyle{\pm0.02}$\\
              \midrule
20 deg.         & S3-Axis       & $100$       & $6.20\scriptstyle{\pm2.63}$      & $0.11\scriptstyle{\pm0.051}$\\
              & U3-Axis     & $100$      & $9.50\scriptstyle{\pm0.64}$       & $0.33\scriptstyle{\pm0.09}$\\
              & S. Shear Only          & $80$        & $4.71\scriptstyle{\pm0.19}$        & $0.08\scriptstyle{\pm0.04}$\\
              & Normal Only              & $100$       & $8.44\scriptstyle{\pm4.05}$      & $0.13\scriptstyle{\pm0.10}$\\
              & Proprio. Only           & Fail      & Fail       & Fail\\
              \toprule
\multicolumn{5}{c}{\textbf{Water Bottle - Variable COM}}  \\   
\toprule
0 deg.          & S3-Axis & $100$    & $9.38\scriptstyle{\pm1.61}$      & $0.08\scriptstyle{\pm0.07}$\\
              & U3-Axis    & $100$     & $8.13\scriptstyle{\pm0.83}$       & $0.27\scriptstyle{\pm0.02}$\\
              & S. Shear Only          & $100$      & $5.57\scriptstyle{\pm4.70}$       & $0.08\scriptstyle{\pm0.11}$\\
              & Normal Only             & $100$      & $8.17\scriptstyle{\pm0.33}$      & $0.1\scriptstyle{\pm0.02}$\\
              & Proprio. Only & $100$        & $10.76\scriptstyle{\pm0.64}$       & $0.13\scriptstyle{\pm0.08}$\\
              \midrule
5 deg.         & S3-Axis     & $100$       & $6.20\scriptstyle{\pm2.63}$      & $0.11\scriptstyle{\pm0.051}$\\
              & U3-Axis     & $100$      & $9.50\scriptstyle{\pm0.64}$       & $0.33\scriptstyle{\pm0.09}$\\
              & S. Shear Only          & $80$        & $4.71\scriptstyle{\pm0.19}$        & $0.08\scriptstyle{\pm0.04}$\\
              & Normal Only              & $100$       & $8.44\scriptstyle{\pm4.05}$      & $0.13\scriptstyle{\pm0.10}$\\
              & Proprio. Only          & Fail      & Fail      & Fail\\
              \midrule
10 deg.        & S3-Axis      & $100$     & $1.36\scriptstyle{\pm1.63}$          & $0.03\scriptstyle{\pm0.03}$\\
              & U3-Axis     & Fail       & Fail        & Fail\\
              & S. Shear Only          & Fail        & Fail       & Fail\\
              & Normal Only             & Fail      & Fail        & Fail\\
              & Proprio. Only          & Fail       & Fail     & Fail \\
              \bottomrule
\end{tabular*}
\caption{\normalfont \small All policy adaptation experimental results. Success out of 5 trials, distance and velocity averaged over successful trials. We are interested in the limits of policy adaptation, so experiments focus on the maximum angles at which policies are still capable of completing the task.}
\label{table:results}
\vspace{-2em}
\end{table}

An overview of our deployment pipeline is shown in Figure \ref{fig:system}, and notably, \textit{all evaluations are conducted in the real world}.

We first test in-domain policy performance on the canonical cylinder object, to compare Proprio-Only and a 3-axis tactile policy. Then, with the same cylinder, we test policy adaptation to OOD hand tilt angles \textit{against} gravity (Figure \ref{fig:avg_adapt}B). These tilt angles passively bias object motion opposing the desired translation direction, thus requiring more force from fingers and gait adaptation to complete the task. This experiment isolates the effect of tilting the hand. Finally, we test policy adaptation to OOD objects \textit{and} hand tilt angles (Figure\ref{fig:avg_adapt}C, Table \ref{table:results}). 
\subsection{Tactile Sensing Boosts In-Domain Task Performance and Adaptation to OOD Hand Tilt}

First, we evaluate the performance of S3-Axis and Proprio-Only policies with the same setting as in training: translate a cylinder with no hand tilt (Figure \ref{fig:avg_adapt}A). We deploy 5 real-world rollouts for each policy. The S3-Axis policy achieves an increase of 38\% translation distance and 94\% greater object velocity compared to the Proprio-Only policy. Additionally, the S3-Axis policy performs more consistently, with lower standard deviations for both task metrics.

Next, we test the effect of hand tilt, an OOD condition that increases the object's tangential force opposing the desired direction of translation. We test hand tilts from 0-15 degrees in increments of 5 degrees. As shown in Figure \ref{fig:avg_adapt}B, the translation distance and object velocity generally decrease as the tilt angle increases. This indicates that task difficulty increases with hand tilt. The finger gait adaptations are less effective and the policy is slower to find effective adaptations to OOD physics. However, S3-Axis still outperforms Proprio-Only in both task metrics, across all tilt angles.

\subsection{Three-Axis Tactile Sensing Enables Superior Out-of-Domain Adaptation}

\begin{figure}[!t]
\centering
\includegraphics[width=\columnwidth]{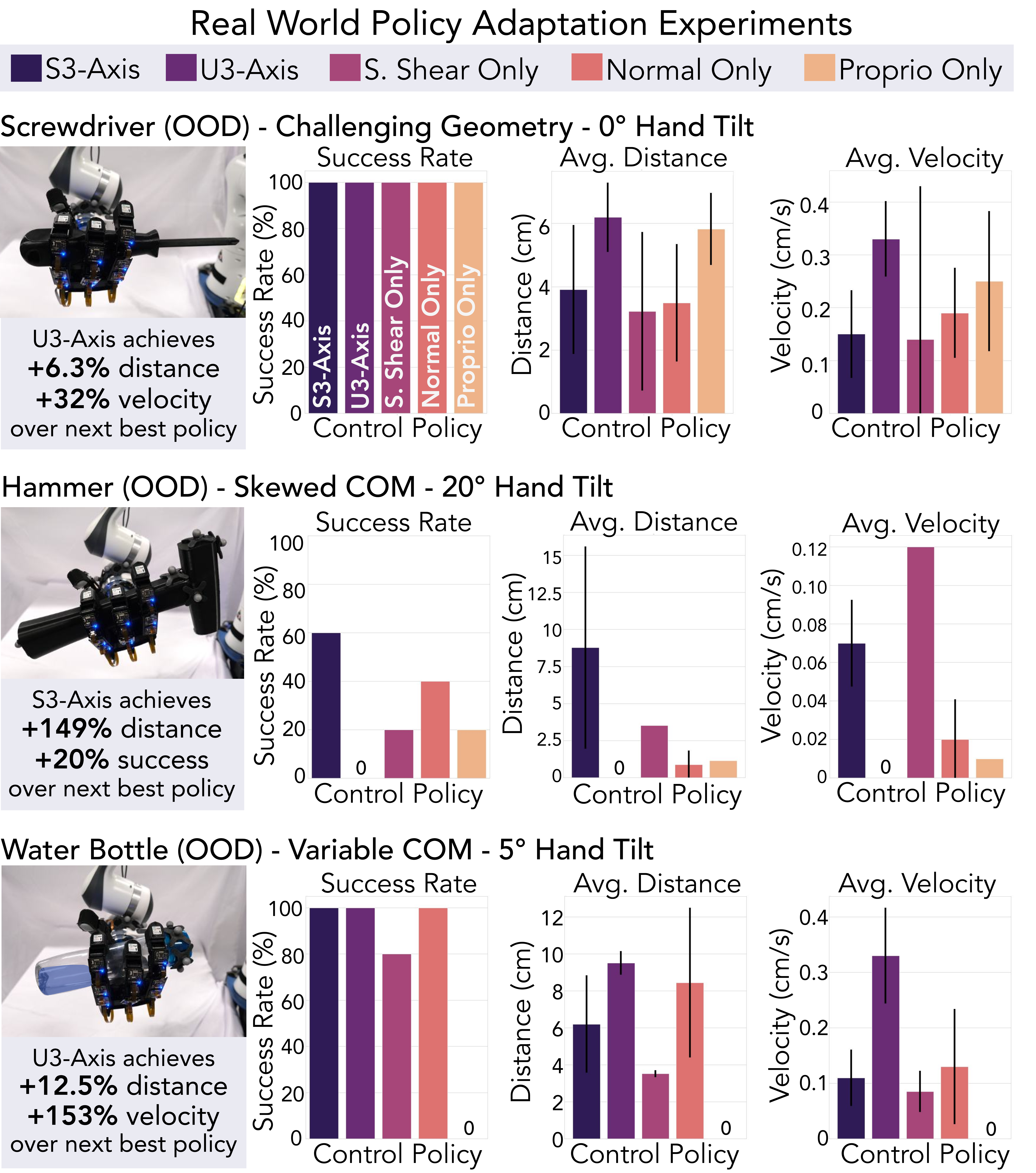}
\vspace{-1.5em}
\caption{\small{\textbf{Highlighted adaptation experiments.} We highlight experiments near policy adaptation limits, and show that our S3-Axis and U3-Axis policies generally achieve the best task metrics. Water in the bottle is annotated for contrast.}}
\label{fig:adapt_exp}
\vspace{-1.9em}
\end{figure}

In this section, we simultaneously test policy adaptation to two classes of perturbations: hand angle tilt \textit{and} OOD objects. The three-axis tactile sensing policies achieve the best policy adaptation. We benchmark our results against Signed Shear Only, Normal Only, and Proprio-Only. In the remaining section and Figure \ref{fig:avg_adapt}B, we show that across all OOD experiments, S3-Axis and U3-Axis policies achieve, on average, superior performance compared to the other baselines. In Figure~\ref{fig:adapt_exp}, we highlight interesting cases.

\textbf{Screwdriver - Challenging Geometry - No Hand Tilt.} We evaluate our policy on the challenging screwdriver object. The screwdriver is highly nonuniform, featuring distinctly discontinuous surfaces that create multiple points of contact with the palm. This leads to very different tactile signatures compared to the training distribution. All policies are capable of the task, but the U3-Axis policy is the most effective, achieving +6.3\% improvement in distance and +32\% improvement in velocity over the next best policy. 

\textbf{Hammer - Skewed COM and Rectangular Geometry - 15-20 Degree Hand Tilt.} The hammer has a skewed COM, located around the intersection of the head and handle. This introduces OOD physics, as torque from gravity applied at the COM (unsupported by the palm) makes the hammer inherently less stable during translation. Also, the handle is rectangular, so there is more contact with the palm, compared to cylinders from training. All policies are capable at 15 degrees, but there is significant performance degradation at 20 degrees. Here, \textit{signed} shear is particularly valuable: the S3-Axis and Signed Shear Only policies achieve the best task metrics. The S3-Axis policy achieves +149\% distance and +20\% success over the next best policy. Surprisingly, the U3-Axis policy completely failed with this case, and the other policies struggled to manipulate the hammer. The failure modes of the policies are: 1) the policy does not adapt to produce an effective gait within 120 s, or 2) the hammer slips and falls.

\textbf{Water Bottle - Variable COM and Irregular Geometry - 0-10 Degree Hand Tilt.} The water bottle has a variable COM as the water sloshes in the bottle and the bottle geometry is an irregular cylinder. When the hand is tilted and at the beginning of the task, the water is at the bottom of the bottle, which applies a torque on that end. As the water bottle moves across, the COM shifts to the center and eventually to the other end. The most common failure mode is when the policy fails to adapt the finger gait within 120 s. The U3-Axis policy achieves +12.5\% distance and +153\% velocity over the next best policy.
\secv
\subsection{Experimental Analysis}
\begin{figure}[!b]
\centering
\includegraphics[width=\columnwidth]{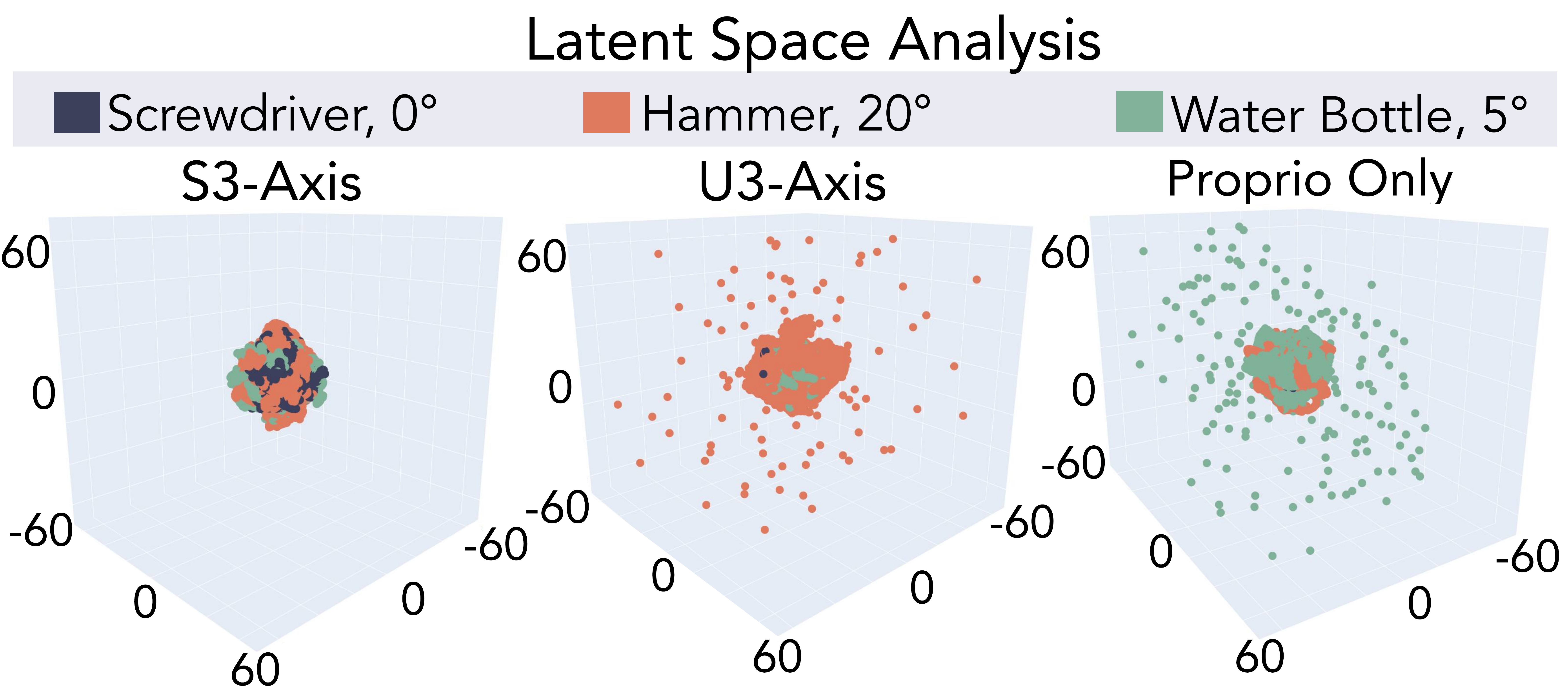}
\vspace{-1.1em}
\caption{\small{\textbf{Latent space analysis.} We use t-SNE to examine the extrinsics vector $\hat{z_t}$ from our experiments in A. Complete policy failure correlates with high dispersion of $\hat{z_t}$, as shown with U3-Axis and the hammer, and Proprio Only with the water bottle.}}
\label{fig:latent_space_analysis}
\vspace{-1em}
\end{figure}
\textbf{Latent Space Analysis.} We analyze the extrinsics vector $\hat{z_t}$ from all 15 rollouts per policy (5 per object) from our experiments in Figure \ref{fig:adapt_exp} with t-SNE to produce the plots in Figure \ref{fig:latent_space_analysis}. We find a correlation between high dispersion of $\hat{z_t}$ clusters and complete policy failure. U3-Axis failed with the hammer at 20 degrees and Proprio Only failed with the water bottle at 5 degrees. The $\hat{z_t}$ clusters for these failed rollouts are more dispersed compared to more successful rollouts. In contrast, the S3-Axis policy was more successful overall and exhibited a dense cluster of $\hat{z_t}$ from all rollouts.

\begin{figure}[!t]
\centering
\includegraphics[width=\columnwidth]{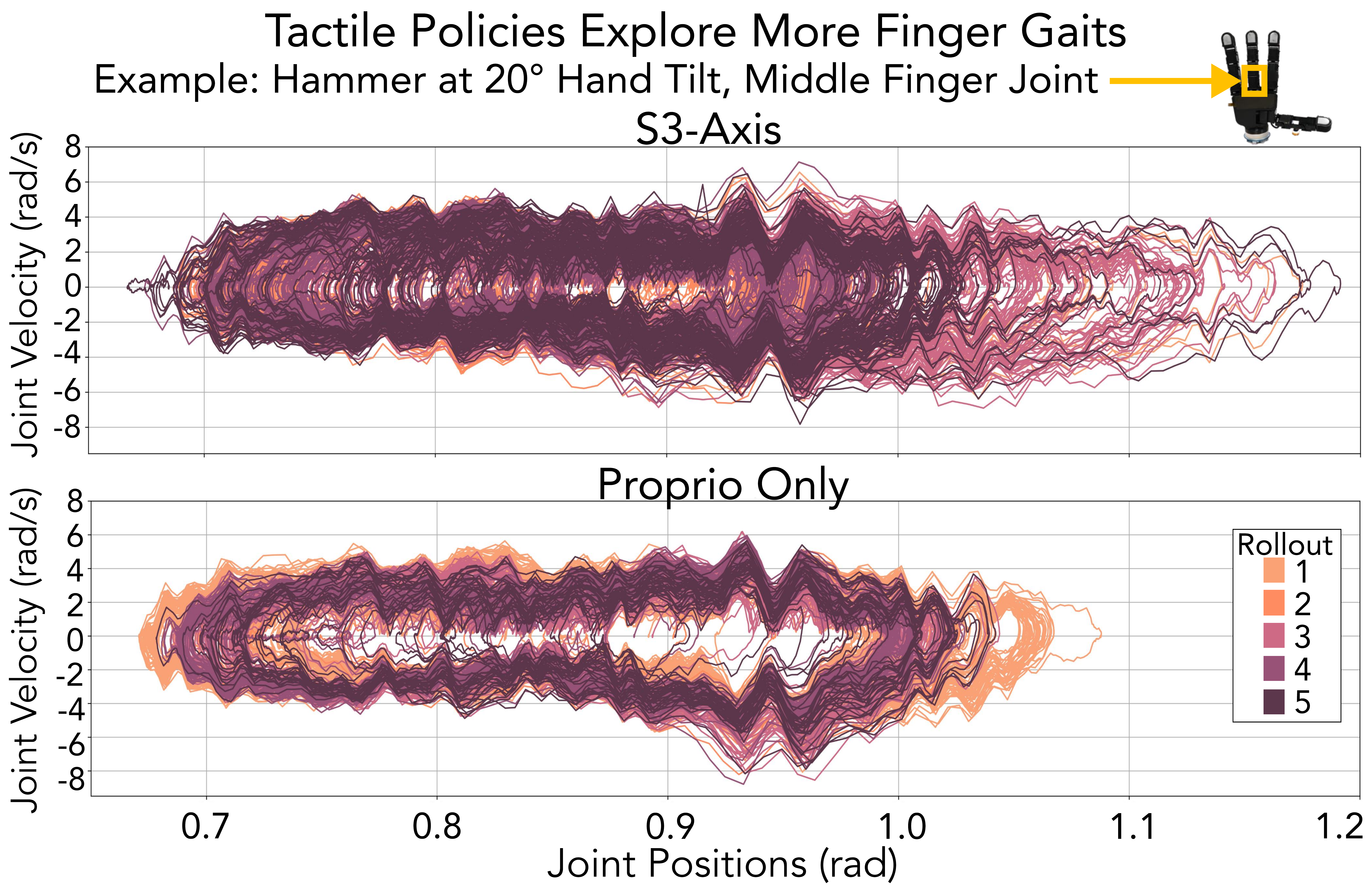}
\vspace{-2em}
\caption{\small{\textbf{Tactile policies explore more finger gaits.} These phase portraits show how gaits differ between policies. Intersections through the Poincar\'e section in the S3-Axis phase portraits have +35\% standard deviation compared to Proprio Only.}}
\label{fig:finger_gaits}
\vspace{-1.3em}
\end{figure}

\textbf{Tactile Policies Explore More Finger Gaits.} We analyze the joint states of all policies from our experiments in Figure \ref{fig:adapt_exp}. We compare the standard deviation of intersections through Poincar\'e sections of the phase portraits for each finger motor and find that on average, tactile policies explore more joint states relative to Proprio Only, particularly for the hammer and water bottle. We highlight the contrast in joint exploration in phase portraits of S3-Axis and Proprio-Only, where S3-Axis has an increase of 35\% standard deviation of intersections through the Poinca\'re section (Figure \ref{fig:finger_gaits}). Greater joint state exploration is potentially a factor in task success and an indicator of gait adaptation; however, other variables such as gait cycle timing and finger coordination are also important to consider. 

\subsection{Demonstrations}

\textbf{Diverse Objects.} We demonstrate our S3-Axis and U3-Axis policies with an additional 15 objects (Supp. Video). Non-cylindrical objects include the mustard bottle, bleach bottle, and multimeter. The object weights range from $57$ g - $524$ g. We also demonstrate with varied object compliances: flexible (cheetos, cups), soft (paper towel roll, pool noodle, soft tripod case), and rigid (80-20 aluminum bar, air can). 

\textbf{Real-World Videos.} Our \href{\web}{Project Website} hosts videos of our experiments, which can be used to observe task metrics and policy gait adaptation.

\section{Conclusion and Future Work}
In this work, we propose a novel sensor model and show it can be used to train RL policies in simulation for a challenging task. Our dexterous in-hand translation policies with ReSkin can be zero-shot deployed to the real-world. We show that ReSkin shear and normal force sensing consistently enables the best ID performance \textit{and} adaptation to both OOD hand orientations and objects. We see these contributions as a key step towards enabling tactile feedback for general in-hand manipulation.

Our current tactile model is heuristic and empirically formulated to match real ReSkin signals. Future work will explore models more aligned with physics, especially for continuous tactile signals. A key challenge in sim-to-real transfer of ReSkin coverage on both the fingers and palm will be modeling cross-talk between multiple discrete sensors. We can also explore domain adaptation methods for sim-to-real tactile transfer, and investigate how 3-axis tactile sensing improves policy performance for other dexterous skills. 


\section*{Acknowledgement}
\vspace{-0.3em}
We thank Mustafa Mukadam, Mike Lambeta, Tingfan Wu, Luis Pineda, Taosha Fan, Patrick Lancaster, Mrinal Kalakrishnan, Raunaq Bhirangi, Carolina Higuera, Akash Sharma, Suddhu Suresh, and William Yang for helpful discussions and feedback throughout this work. We thank Dr. Nadia Figueroa, Ho Jin Choi, Tianyu Li, and Harshil Parekh for assistance with the Franka and Optitrack system.


\bibliographystyle{IEEEtran}
\bibliography{main}

\end{document}